\documentclass{article} 
\usepackage{iclr2016_conference,times}
\usepackage{url}
\usepackage{appendix}

\usepackage{csquotes}

\usepackage[utf8]{inputenc}
\usepackage{natbib}

\newcommand{\parencite}[1]{\citep{#1}}
\newcommand{\textcite}[1]{\cite{#1}}

\usepackage{amssymb}
\usepackage{amsmath }
\usepackage{units}
\usepackage{times}
\usepackage{graphics}
\usepackage{graphicx}
\usepackage{algorithm}
\usepackage{algorithmicx}
\usepackage[noend]{algpseudocode}
\usepackage{color}
\algnewcommand{\LineComment}[1]{\State \(//\) #1}

\title{Online Semi-Supervised Learning with Deep Hybrid Boltzmann Machines and Denoising Autoencoders}

\author{
Alexander G. Ororbia II \& C. Lee Giles \& David Reitter
\\
College of Information Science \& Technology\\
The Pennsylvania State University\\
University Park, PA 16801, USA \\
\texttt{\{ago109,giles,reitter\}@psu.edu} \\
}

\begin{document}

\maketitle

\begin{abstract}


Two novel deep hybrid architectures, the Deep Hybrid Boltzmann Machine and the Deep Hybrid Denoising Auto-encoder, are proposed for handling semi-supervised learning problems.  The models combine experts that model relevant distributions at different levels of abstraction to improve overall predictive performance on discriminative tasks.   Theoretical motivations and algorithms for joint learning for each are presented. We apply the new models to the domain of data-streams in work towards life-long learning. The proposed architectures show improved performance compared to a pseudo-labeled, drop-out rectifier network. 
\end{abstract}

\section{Introduction}
\label{intro}
Unsupervised pre-training can help construct an architecture composed of many layers of feature detectors \parencite{erhan_why_2010}.  Even though the ultimate task is discriminative, a generative architecture, such as the Deep Belief Network (DBN) or Stacked Denoising Autoencoder (SDA), may be first used to initialize the parameters of a multi-layer perceptron (MLP) which then is fine-tuned to the supervised learning task \parencite{bengio_greedy_2007}.  

However, learning the parameters of the unsupervised architecture is quite difficult, often with little to no grasp of the final influence the generative parameters will have on the final discriminative model \parencite{larochelle_learning_2012,goodfellow_multi-prediction_2013}.  These architectures often feature many hyper-parameters that affect generalization performance, quickly creating a challenging tuning problem for human users.
Furthermore, though efficient, the generative models used for pre-training learnt greedily carry the potential disadvantage of not providing ``global coordination between the different levels'' \parencite{bengio_how_2014}, the sub-optimality of which was empirically shown in \textcite{arnold_layer-wise_2012}. This issue was further discussed as the problem of ``shifting representations'' in \textcite{ororbia_deep_hybrid_2015a}, where upper layers of a multi-level model are updated using immature latent representations from layers below potentially leading to unstable learning behavior or worsened generalization performance. While greedily built models can be later tuned jointly to construct architectures such as the Deep Boltzmann Machine (DBM, \cite{salakhutdinov_deep_2009}) or via the Wake-Sleep algorithm \parencite{hinton_wake_sleep_1995} (and improved variants thereof \parencite{bornschein_reweighted_2014}) the original training difficulty remains.


One way to exploit the power of representation learning without the  difficulties of pre-training is to instead solve the hybrid learning problem:  force a model to balance multiple supervised and unsupervised learning objectives in a principled manner.
Many recent examples demonstrate the power and flexibility of this approach \parencite{larochelle_classify_2008,ranzato_semi-supervised_2008,socher_semi_supervised_2011,larochelle_learning_2012,ororbia_deep_hybrid_2015a,ororbia_deep_hybrid_2015b}. The Stacked Boltzmann Expert Network (SBEN) and its autoencoder variant, the Hybrid Stacked Denoising Autoencoder (HSDA, \cite{ororbia_deep_hybrid_2015a}) were proposed as semi-supervised deep architectures that combined the expressiveness afforded by a multi-layer composition of non-linearities with a more practical approach to model construction.  

Though promising, the previous approaches for learning deep hybrid architectures still suffer from some key issues:  1) parameters are learnt in a layer-wise fashion, which means that these models are susceptible to the ``shifting representations'' problem, and 2) the predictive potential of hybrid architectures' multiple layers has previously involved a naive form of vertical aggregation, whereas a more principled unified approach to layer-wise aggregation could lead to further performance improvements. In this paper, we propose two novel architectures and a general learning framework to directly address these problems while still providing an effective means of semi-supervised learning.

The problem of semi-supervised online learning is modeled after a scientific inquiry: how do babies learn?  They start from representational beginnings that could range from blank slate to specific computational constraints.  They learn from a  stream of observations by perceiving the environment and (generatively) by trying to interact with it.  While there are plenty of observations, the majority is unlabeled;
the feedback they receive on their interpretations is uninformative and has been claimed to be too poor to facilitate learning without extensive priors \parencite{chomsky1980rules}.  It seems obvious that online semi-supervised learning is a key task to achieving artificial general intelligence.

\section{The Multi-Level Semi-supervised Hypothesis}
\label{semi_sup_hypoth}
Our motivation for developing hybrid models comes from the semi-supervised learning (prior) hypothesis of \textcite{rifai_manifold_2011}, where learning aspects of $p(\mathbf{x})$ improves the model's conditional $p(y|\mathbf{x})$.  The hope is that so long as there is some relationship between $p(\mathbf{x})$ and $p(y|\mathbf{x})$, a learner may be able to make use of information afforded by cheaply obtained unlabeled samples in tandem with expensive labeled ones. 

\textcite{ororbia_deep_hybrid_2015a,ororbia_deep_hybrid_2015b} showed that a hybrid neural architecture, $L$ layers deep, could combine this hypothesis with the expressiveness afforded by depth.  Each layer-wise expert (or building block model) could be used to compute $p(y|\mathbf{h}_l)$ for $l = [0, L]$ and vertically aggregated to yield improved predictive performance.  The Hybrid Stacked Denoising Autoencoders (\emph{HSDA}) model, a stack of single-layer MLP coupled with single-layer auto-associators, directly embodies the multi-level view of the semi-supervised learning prior hypothesis, or rather, what we call the ``weak multi-level semi-supervised learning hypothesis''~\footnote{``Weak'' refers to the possibility that $p(\mathbf{x})$ may be loosely related to $p(y|\mathbf{x})$, if at all.  Learning $p(\mathbf{x})$ may or may not help with prediction.}.  According to this hypothesis, for an architecture designed to learn \emph{L} levels of abstraction of data, learning something about the marginal $p(\mathbf{h}_l)$ along with $p(y|\mathbf{h}_l)$ for $l = [0, L]$, will improve predictive performance on $p(y|\mathbf{x})$.  The Deep Hybrid Denoising Autoencoder, our proposed joint version of the HSDA presented in Section \ref{dhda}, also embodies the ``weak multi-level semi-supervised learning hypothesis''.

An alternative model is the Stacked Boltzmann Experts Network (\emph{SBEN}) model, a stack of hybrid restricted Boltzmann machines, where instead each layer-wise expert attempts to model a joint distribution at level \emph{l}, $p(y, \mathbf{h}_l)$.  In the \emph{SBEN}, aggregating the resulting $p(y|\mathbf{h}_l)$ for $l = [0, L]$, can ultimately improve predictive performance on $p(y|\mathbf{x})$.  This we call the ``strong multi-level semi-supervised learning hypothesis''~\footnote{``Strong'' refers to the fact we know $p(y|\mathbf{x})$ is related to $p(y,\mathbf{x})$ (i.e., $p(y|\mathbf{x}) = p(y,\mathbf{x})/p(\mathbf{x})$).  Thus learning the joint will yield information relevant to the conditional.}.  The DHBM, our proposed joint version of the SBEN presented in Section \ref{dhbm}, likewise embodies this hypothesis.

Some experimental evidence has been provided to support these two variant hypotheses.  In the work of \cite{ororbia_deep_hybrid_2015a}, a 3-layer SBEN outperformed a semi-supervised deep rectifier network (among other base-lines) by nearly 14\% in image classification tasks.  In \cite{ororbia_deep_hybrid_2015b} a 3-layer SBEN (trained via the bottom-up-top-down algorithm) was shown to outperform a semi-supervised rectifier network and the original SBEN of \cite{ororbia_deep_hybrid_2015a} (and other competitive text classifiers) consistently in text categorization experiments by as much as 36\%. The works of \textcite{calandra_learning_2012,zhou_online_2012}, which are special cases of the HSDA, are also further experimental evidence that support the weak multi-level semi-supervised hypothesis. 

\section{Deep Hybrid Model Architectures}
\label{arch}

Below, we present the design of two candidates for building unified, deep hybrid architectures, namely, the Deep Hybrid Boltzmann Machine (DHBM) and the Deep Hybrid Denoising Auto-encoder (DHDA).  

\subsection{The Deep Hybrid Boltzmann Machine}
\label{dhbm}

\begin{figure}[!t]
\centering
\includegraphics[width=3.0in]{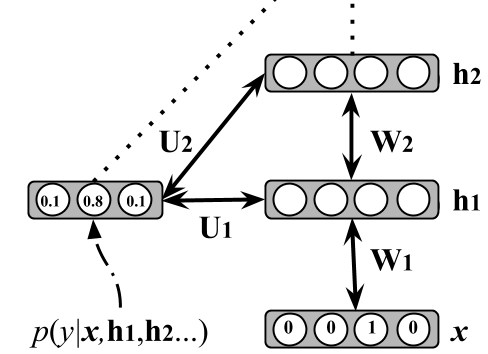}
\caption{The full deep hybrid Boltzmann machine architecture.  Note the bi-directional nature of the weights, meaning most sets of variables require both a bottom-up \& top-down calculation.}
\label{dhbm_arch}
\end{figure}  

Like the Deep Boltzmann Machine (DBM) is to the Deep Belief Network (DBN), the DHBM can be viewed as a more sophisticated version of an SBEN.  The primary advantage of a DHBM over the DBM, much like that of the SBEN over the DBN, is that the hybrid architecture is learnt with the ultimate intent of performing classification.  This entails tracking model performance via classification error or some discriminative loss and avoids the need for expensive, non-trivial methods as annealed importance sampling (AIS, \cite{neal_annealed_2001}) to estimate partition functions for approximate objectives \parencite{hinton_training_2002}.

Instead of a single model, another way to view a DHBM is simply as a composition of tightly integrated hybrid restricted Boltzmann machines (HRBM's).  An SBEN is essentially a stack of HRBM's that each models $p(y,\mathbf{h}^l)$ at their own respective level $l$ of abstraction in the overall architecture. To exploit predictive power at each level of abstraction, one applies an averaging step of all layer-wise predictors to compute the SBEN's $p(y|\mathbf{x})_{ensemble}$ at inference time.  The initial version of this model was trained using a greedy, bottom-up approach, where each layer learned to predict independently of the outputs of other layers (but was conditioned on the latent representation of the layer below).  \textcite{ororbia_deep_hybrid_2015b} introduced a degree of joint learning of SBEN parameters through the \emph{Bottom-Up-Top-Down} algorithm to improve performance, however, its bottom-up generative gradient step was still layer-wise in nature.

Using the comparative view of the SBEN, one creates a DHBM by taking a stack of HRBM experts and couples their predictors together, which means the overall model's prediction is immediately dependent on what all layers have learnt with respect to their level of abstraction.  Furthermore, one makes the connections between layer-wise experts fully bi-directional, meaning that in order compute the state of any latent variable layer in the model (except for the input and top-most latent layers), one needs to incorporate activations from both the layer immediately below and the layer immediately above.  
A simple depiction of the first 3 layers of such a model is depicted in Figure~\ref{dhbm_arch}.  As a result, we have built a plausible machine that jointly leverages its multiple levels of abstraction to model the joint distribution of labeled data as opposed to a simple stack of joint distribution models that characterize the SBEN.

With the above picture in mind, we may explicitly define the 3-layer DHBM (or \emph{3}-DHBM), though as shown in Figure~\ref{dhbm_arch} this definition extends to an \emph{L}-layer model.  With pattern vector input $\textbf{x} = (x_1,\cdots,x_D)$ and its corresponding target variable $y \in \{1,\cdots,C\}$, utilizing two sets of latent variables $\mathbf{h}^1 = (h^1_1,\cdots,h^1_{H_1})$ and $\mathbf{h}^2 = (h^2_1,\cdots,h^2_{H_2})$ and model parameters $\Theta^m = (\mathbf{W}^1, \mathbf{U}^1, \mathbf{W}^2, \mathbf{U}^2)$\footnote{Note that we omit hidden and visible bias terms for simplicity.}, the energy of a DHBM is:

\begin{equation}
\label{dhbm_energy}
E(y,\mathbf{x},\mathbf{h}^1,\mathbf{h}^2) = -\mathbf{h}^{1\top}\mathbf{W}^1\mathbf{x} - \mathbf{h}^{1\top}\mathbf{U}^1\mathbf{e}_y -\mathbf{h}^{2\top}\mathbf{W}^2\mathbf{h}^1 - \mathbf{h}^{2\top}\mathbf{U}^2\mathbf{e}_y \mbox{.}
\end{equation}

\noindent
where we note that \(\textbf{e}_y = (\mathbf{1}_{i=y})^C_{i=1} \) is the one-hot vector encoding of $y$. The probability that the \emph{3}-DHBM assigns to the 4-tuple $(y,\mathbf{x},\mathbf{h}^1,\mathbf{h}^2)$ is:

\begin{equation}
\label{prob}
p(y,\mathbf{x},\Theta) = \frac{1}{Z} \sum_{\mathbf{h}} e^{(E(y,\mathbf{x},\mathbf{h}^1,\mathbf{h}^2))}
\end{equation}

\noindent
where $Z$ is the partition function meant to ensure a valid probability distribution (calculated by summing over all possible model configurations).

Noting the introduction of top-down calculations, the visible and latent states of the 3-DHBM may be computed via the following implementable equations:

\begin{equation}
\label{sample_h1_given_y_x_h2}
p(\textbf{h}^1|y,\textbf{x},\textbf{h}^2) = \prod_{j} p(h^1_j|y,\textbf{x},\textbf{h}^2),\mbox{ with } p(h^1_j = 1|y,\textbf{x}) = \phi(U^1_{jy} + \sum_{i} W^1_{ji}x_i + \sum_{k} W^2_{kj}h^2_k)
\end{equation}

\begin{equation}
\label{sample_h2_given_y_h1}
p(\textbf{h}^2|y,\textbf{h}^1) = \prod_{k} p(h^2_k|y,\textbf{h}^1),\mbox{ with } p(h^2_k = 1|\textbf{h}^1) = \phi(U^2_{ky} +\sum_{j} W^2_{kj}h^1_j)
\end{equation}

\begin{equation}
\label{sample_x_given_h}
p(\textbf{x}|\textbf{h}^1) = \prod_{i} p(x_i|\textbf{h}^1),\mbox{ with } p(x_i = 1|\textbf{h}) = \phi(\sum_{j} W^1_{ji}h^1_j)
\end{equation}

\begin{equation}
\label{sample_y_given_h1_h2}
p(y|\textbf{h}^1,\textbf{h}^2) = \frac{e^{\sum_{j} U^1_{jy}h^1_j + \sum_{j} U^2_{jy}h^2_j}}{\sum_{y^{\star}} e^{\sum_{j} U^1_{jy^{\star}}h^1_j + \sum_{j} U^2_{jy^{\star}}h^2_j}}
\end{equation}

\noindent
where the activation $\phi(v) = {1}/({1 + e^{-v}})$, or the logistic sigmoid, and $y$ is used to access a particular class filter from $U^l$.  In the interest of adapting the model to different types of input, such as continuous-valued variables, $\phi(v)$ itself can be switched to alternative functions such as the rectified linear unit.
One may use this set of equations as the fixed-point formulas for running mean-field inference in the deep architecture to obtain $\mu = \{\mu^1,\mu^2\}$.
More importantly, one may notice the dependency between these conditionals and as a result one may use a bottom-up pass approximation with weight doubling to initialize the mean-field like that of \textcite{salakhutdinov_deep_2009}.  One must then run several additional steps of mean-field, cycling through Equations \ref{sample_h1_given_y_x_h2}, \ref{sample_h2_given_y_h1}, \ref{sample_x_given_h}, and \ref{sample_y_given_h1_h2}, to get the model's reconstruction of the input and target (or model prediction).

To speed up both training and prediction time (since the goal is to make use of a single bottom-up pass), we propose augmenting the DHBM architecture with a co-model, or separate auxiliary network, which was previously utilized to infer the states of latent variables in the DBM for a single bottom-up pass~\parencite{salakhutdinov_efficient_2010}.  The recognition network, or MLP serving the role of function approximation, can be effectively fused with the deep architecture of interest and trained via a gradient descent procedure.  Underlying the co-training of a separate recognition network is the expectation that the target model's mean-field parameters will not change much after only a single step of learning, which was also empirically shown in \textcite{salakhutdinov_efficient_2010} to work in practically training a DBM.  The same principle we claim holds for a deep hybrid architecture, such as the DHBM, trained in a similar fashion. 

The recognition network, the weights of which are initialized to those of the DHBM at the start of training, is specifically tasked with computing a fully factorized, approximate posterior distribution as shown below:

\begin{equation}
\label{g_rec}
	Q^{rec}(\mathbf{h}|\mathbf{v};\textbf{\textit{v}}) = \prod^{H_1}_{j=1} \prod^{H_2}_{k=1} q^{rec}(h^1_j)q^{rec}(h^2_k)
\end{equation}

\noindent
where the probability $q^{rec}(h^l_i = 1) = v^l_i$ for layers $l = 1,2$ and $\upsilon = \{\mathbf{v}^1,\mathbf{v}^2\}$. Running the recognition network, with parameters $\Theta^{rec} = (R^1,R^2)$ (again, omitting bias terms for simplicity), is straightforward, as indicated by the equations below that constitute its feedforward operation:

\begin{minipage}{0.45\linewidth}
\begin{equation}
\label{g_rec_op1}
	v^1_j = \phi(\sum^D_{i=1} 2R^1_{ij}v_i)
\end{equation}

\end{minipage} \hspace{1cm} 
\begin{minipage}{0.45\linewidth}

\begin{equation}
\label{g_rec_op2}
	v^2_j = \phi(\sum^{H_1}_{j=1} R^2_{jk}v^1_j)
\end{equation}
\end{minipage}

\noindent
where the inference network's weights are doubled at each layer (except the top layer) to compensate for missing top-down feedback. Note that after initialization, the weights of the inference network are no longer shared with the original model. The inference network can be used to reasonably guess the values of the fixed-point mean-field to compute the values for $(\mathbf{h}^1, \mathbf{h}^2)$ and then Equations \ref{sample_h1_given_y_x_h2}, \ref{sample_h2_given_y_h1}  \ref{sample_x_given_h}, and \ref{sample_y_given_h1_h2} may be run for a subsequent single mean-field step.  More importantly, in our hybrid model definition, during prediction time, the architecture may directly generate the appropriate prediction for $y$ by using the trained recognition network to infer the latent states of the DHBM.

The recognition network is trained according to the following objective:

\begin{equation}
\label{recog_net_loss}
	KL(Q^{MF}(\mathbf{h}|\mathbf{v};\mu)||Q^{rec}(\mathbf{h}|\textbf{v};\upsilon)) = -\sum_i \mu_i log v_i -\sum_i (1 - \mu_i) log(1 - \upsilon_i) + Const
\end{equation}

\noindent
which is the minimization of the Kullback-Leibler (KL) divergence between $Q^{MF}(\mathbf{h}|\mathbf{v};\mathbf{\mu})$, the posterior of the DBM mean-field, and $Q^{rec}(\mathbf{h}|\textbf{v};\textit{v})$, the factorial posterior of the recognition network.


\subsection{The Deep Hybrid Denoising Autoencoder}
\label{dhda}

Following a similar path as the previous section but starting from the HSDA base architecture, we also propose the autoencoder variant of the DHBM, the DHDA.  This also borrows the same underlying model structures of the DHBM, including the bi-directional connections needed for incorporating top-down and bottom-up influence.  However, instead of learning via a Boltzmann-based approach, we shall learn a stochastic encoding and decoding process through multiple layers jointly.  The DHDA may alternatively viewed as a stack of tightly integrated hybrid denoising autoencoder (HdA) building blocks with coupled predictors. An HdA is a single hidden-layer MLP that shares its input-to-hidden weights with an encoder-decoder model whose own weights are tied (i.e., decoding weights are equal to the transpose of the encoding weights).

A 3-layer version of the joint model (also generalizable to \emph{L} layers and defined by the same parameter-set as the DHBM) is specified by the following set of encoding ($f^1_{\theta}(y,\widehat{\mathbf{x}},\widehat{\mathbf{h}^2}),f^2_{\theta}(\widehat{\mathbf{h}^1})\}$) and decoding ($g_{\theta}(\widehat{\mathbf{h}^1})$) functions:

\begin{equation}
\label{encode_h1_given_x_h2}
\mathbf{h}^1 = f^1_{\theta}(\widehat{\mathbf{x}},\widehat{\mathbf{h}^2}) = \phi(W^1 \widehat{\mathbf{x}} + (W^2)^T \widehat{\mathbf{h}}^2)
\end{equation}

\begin{minipage}{0.475\linewidth}
\begin{equation}
\label{encode_h2_given_h1}
\textbf{h}^2 = f^2_{\theta}(\widehat{\mathbf{h}^1}) = \phi(W^2 \widehat{\mathbf{h}}^1)
\end{equation}

\end{minipage} \hspace{0.2cm} 
\begin{minipage}{0.475\linewidth}

\begin{equation}
\label{decode_x_given_h}
\bar{\mathbf{x}} = g_{\theta}(\widehat{\mathbf{h}^1}) = \phi((W^1)^T \widehat{\mathbf{h}}^1)
\end{equation}
\end{minipage}

\noindent
where $\{W^1,W^2\}$ are weight matrices connecting input $\mathbf{x}$ to $\mathbf{h}^1$ and $\mathbf{h}^1$ to $\mathbf{h}^2$ respectively (the superscript \emph{T} denotes a matrix transpose operation). The output function $o_{\theta}(y | \mathbf{h}^1,\mathbf{h}^2)$ needed to generate predictions is calculated via Equation \ref{sample_y_given_h1_h2} much like the DHBM model.  Like the HSDA, the DHDA uses a stochastic mapping function \(\widehat{\mathbf{h}}^l_t \sim q_{\mathcal{D}}(\widehat{\mathbf{h}}^l_t|\mathbf{h})\) to corrupt input vectors (i.e., randomly masking entries under a given probability)~\footnote{Note that $\mathbf{h}^0 = \mathbf{x}$. While we opted for a simple denoising-based model, one could modify the building blocks of the DHDA to any alternative, such as one that makes uses of a contractive penalty \parencite{rifai_contractive_2011}.}.  Note that the DHDA requires less matrix operations than the DHBM, since it exemplifies the ``weak multi-level semi-supervised hypothesis'', giving it the advantage of a speed-up compared to the DHBM.

To calculate layer-wise activation values for the DHDA, one uses the recognition network to obtain initial guesses for $\{\mathbf{h}^1,\mathbf{h}^2\}$ and then cycles through Equations \ref{encode_h1_given_x_h2}, \ref{encode_h2_given_h1}, \ref{decode_x_given_h}, and Equation \ref{sample_y_given_h1_h2} much like the mean-field procedure of the DHBM.

\begin{figure}[tb]
\centering
\includegraphics[width=0.9\textwidth]{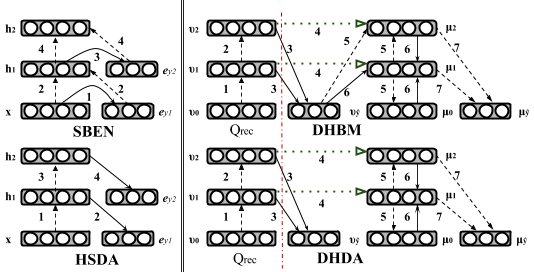}
\label{arch_comp}  
  \caption{The DHBM and DHDA architectures compared side-by-side with their layer-wise predecessors, the SBEN and the HSDA.  The flow of information to gather layer-wise statistics is indicated by the arrows (which represent an operation, such as matrix-multiplication followed by an element-wise non-linearity), with the appropriate layer parameter matrix and the vector at the arrow's origin), which are numbered according to the sequential computational steps taken to calculate them.  Arrows (or operations) in the same computational step are numbered the same and point to the resulting activation values that is to be calculated. $\upsilon$ corresponds to the recognition network's ($Q_{rec}$) initial guess for the mean-field while $\mu$ represents the actual mean-field statistic. Horizontal dotted arrows simply indicate a copy operation.}
\end{figure}

To solidify the comparison between layer-wise and joint models, Figure \ref{arch_comp} shows the DHDA and DHBM architectures and their predecessor non-joint variants.

\subsection{Joint Parameter Learning Methods}
\label{training}

Below, we present the general learning framework for jointly training parameters of a deep hybrid architecture, such as the DHBM and the DHDA described earlier.  Note that since all parameters are jointly modified, the problem of shifting representations is no longer a concern--all layers of the hybrid architecture are globally coordinated during learning. Under this framework, one may employ a variety of estimators for calculating parameter gradients.  In particular we shall describe the two used in this study, namely, stochastic maximum likelihood (for the DHBM) and back-propagation of errors (for the DHDA).  Before proceeding, however, we shall first define the hybrid objective functions that both architectures attempt to minimize.

The DHDA is designed to minimize the following hybrid loss function:

\begin{equation}
\label{dhda_loss}
\mathcal{L}(\mathcal{D}_{lab},\mathcal{D}_{unlab}) = \alpha (-\sum^{|\mathcal{D}_{lab}|}_{t=1} \log p(y | \textbf{x}_t) + \mathcal{L}_{CE}(\mathcal{D}_{lab})) + \beta (-\sum^{|\mathcal{D}_{unlab}|}_{t=1} \log p(y^{\prime} | \textbf{x}_t) + \mathcal{L}_{CE}(\mathcal{D}_{unlab}))
\end{equation}

\noindent
where $y^{\prime}$ is a psuedo-label generated for an unlabeled sample given the current DHDA model and $\mathcal{D}_{lab}$ is the set of labeled samples and $\mathcal{D}_{unlab}$ is the set of unlabeled samples. \(\alpha\) and \(\beta\) are coefficient handles to explicitly control the effects that supervised and unsupervised gradients have on the model's learning procedure, respectively. The loss function $\mathcal{L}_{CE}$ is reconstruction cross entropy, defined as

\begin{equation}
\label{cross_entropy}
\mathcal{L}_{CE}(\mathcal{D}) = -\sum^{|\mathcal{D}|}_{t=1} \mathbf{x}_t \log \mathbf{z}_t + (1 - \mathbf{x}_t) \log (1 - \mathbf{z}_t) \mbox{.}
\end{equation}

The DHBM, on the other hand, minimizes the following hybrid objective:

\begin{equation}
\label{dhbm_loss}
\mathcal{L}(\mathcal{D}_{lab},\mathcal{D}_{unlab}) = -\alpha \sum^{|\mathcal{D}_{lab}|}_{t=1} \log p(y_t,\textbf{x}_t) -\beta \sum^{|\mathcal{D}_{unlab}|}_{t=1} \log p(\textbf{x}_t) \mbox{.}\footnote{We note that the DHBM's hybrid loss function could be augmented with a pair of direct discriminative gradients that additionally minimize $-\sum^{|\mathcal{D}|}_{t=1} \log p(y | \textbf{x}_t)$ for both $\mathcal{D}_{lab}$ and $\mathcal{D}_{unlab}$ using the ensemble back-propagation algorithm proposed in \cite{ororbia_deep_hybrid_2015b}.  This could further improve performance but we found not necessary in this paper's experiments.}
\end{equation}

\noindent
Each hybrid loss corresponds to an appropriate variant of the multi-level semi-supervised prior hypothesis described in Section \ref{semi_sup_hypoth}. The DHDA's loss (Equation \ref{dhda_loss}) embodies the weak variant since it couples a conditional model learning objective with that of a reconstruction model (that deals with the data input directly without labels).  The DHBM's loss (Equation \ref{dhbm_loss}) maps to the strong variant since it directly attempts to model joint densities to yield a useful conditional one. 

\subsubsection{The Joint Hybrid Learning Framework}
\label{learning_framework}

The general learning framework can be decomposed into 3 key steps: 1) gather mean-field statistics via approximate variational inference, 2) compute gradients to adjust the recognition network if one was used to initialize the mean-field, and 3) compute the gradients for the joint hybrid model using a relevant, architecture-specific algorithm.  This procedure works for either labeled or unlabeled samples, however, in the latter case, we either make use of the model's current estimate of class probabilities or a pseudo-label \parencite{lee_pseudo-label:_2013} to create a proxy label.  The full, general procedure is depicted in Algorithm \ref{hybrid_construct}.

In offline learning settings, we note that one could, like in \textcite{salakhutdinov_efficient_2010}, make use of greedy, layer-wise pre-training to initialize a DHBM or DHDA very much in the same way learning a DBN can be used to initialize a DBM.  In fact, learning an \emph{N}-SBEN could be a precursor to learning a DHBM and a learning an \emph{N}-HSDA a precursor to a DHDA, noting that after this first training phase, one would simply tie together the disparate predictor arms of this initial hybrid architecture to formulate the joint predictor needed to calculate Equation \ref{sample_y_given_h1_h2} for either possible joint model.  This form of pre-training would be simpler to monitor, as, like the final target models, accuracy or discriminative loss can be used as a tracking objective as in \textcite{ororbia_deep_hybrid_2015a,ororbia_deep_hybrid_2015b}.


What distinguishes the DHBM from the DHDA, aside from  architectural considerations, is contained in the {\sc calcParamGradients} routine.  The details of each we will briefly describe next and fully explicate in Appendices \ref{appendix_mf} and \ref{appendix_sap}.


\begin{algorithm}[t]
\begin{algorithmic}
\State \textbf{Input:} 1) Labeled $(y,\mathbf{x})$ and unlabeled $(\mathbf{u})$ samples or mini-batches, 2) learning rate \(\lambda\), hyper-parameters \(\beta\), \emph{numSteps} (i.e., \# of mean field steps) and specialized hyper-parameters $\Xi$, and 3) initial model parameters \(\Theta^m = \{\Theta^m_1, \Theta^m_2,...,\Theta^m_N\}\) and recognition model parameters \(\Theta^{rec} = \{\Theta^{rec}_1, \Theta^{rec}_2,...,\Theta^{rec}_L\}\)

\Function{updateModel}{$(y,\mathbf{x})$, $(\mathbf{u})$, $\lambda$,  $\beta$, $numSteps$, $\Xi$, $\Theta^m$, $\Theta^{rec}$}
	\State Use recognition model $\Theta^{rec}$ to calculate $\upsilon$ of approximate factorial posteriors $Q^{rec}_{lab}$ for $(y,\mathbf{x})$ and $Q^{rec}_{unlab}$ for $(\mathbf{u})$ (Eqs. \ref{g_rec_op1}, \ref{g_rec_op2})
    \State Use $Q^{rec}_{unlab}$ to generate a proxy label $\hat{y}$ for $(\mathbf{u})$ via Eq. \ref{sample_y_given_h1_h2}
	\State Set $\mu = \upsilon$ and run mean-field updates (Eqs. \ref{sample_h1_given_y_x_h2}, \ref{sample_h2_given_y_h1}, \ref{sample_x_given_h}, \ref{sample_y_given_h1_h2} or \ref{encode_h1_given_x_h2}, \ref{encode_h2_given_h1}, \ref{decode_x_given_h}, \ref{sample_y_given_h1_h2}) for $numSteps$ to acquire mean-field approximate posteriors $Q^{MF}_{lab}$ and $Q^{MF}_{unlab}$, and mean-field labels $\hat{y}_{lab}$ and $\hat{y}_{unlab}$
    \State Adjust recognition model parameters via one step of gradient descent for both $(y,\mathbf{x})$ and $(\mathbf{u})$, weighting the gradients accordingly:  $\Theta^{rec} \gets \Theta^{rec} - \lambda (\bigtriangledown^{rec}_{lab} + \beta \bigtriangledown^{rec}_{unlab})$, where $\bigtriangledown^{rec}_{lab}$ and $\bigtriangledown^{rec}_{unlab}$ are calculated via back-propagation (with KL-divergence objective)
    \State $\bigtriangledown^{m}_{lab}\gets \Call{calcParamGradients}{(y,\mathbf{x}), \hat{y}_{lab}, Q^{rec}_{lab}, Q^{MF}_{lab}, \Theta^{m}, \Xi}$ 
     \State $\bigtriangledown^{m}_{unlab} \gets \Call{calcParamGradients}{(\hat{y},\mathbf{u}), \hat{y}_{unlab}, Q^{rec}_{unlab}, Q^{MF}_{unlab}, \Theta^{m}, \Xi}$ 
     \State $\Theta^{m} \gets \Theta^{m} + \lambda (\bigtriangledown^{m}_{lab} + \beta \bigtriangledown^{m}_{unlab})$ \Comment{Final update to model parameters}
\EndFunction

\end{algorithmic}
\caption{The general learning framework for performing a single parameter update to an \emph{L}-layer hybrid architecture, where \emph{L} is the desired number of latent variable layers.}
\label{hybrid_construct}
\end{algorithm}

\subsubsection{Mean-Field Contrastive Divergence \& Back-Propagation}
\label{mf_cd}
One simple estimator for calculating the gradients of a hybrid architecture only makes use of two sets of multi-level statistics obtained from the recognition network and running the mean-field equations for a single step.  This results in a set of ``positive phase'' statistics, which result from the data vectors clamped at the input level of the model, and ``negative phase'' statistics, which result from a single step of the model's free-running mode.  One may then use these two sets of statistics to calculate parameter gradients to move the hybrid model towards more desirable optima.  The explicit procedure is presented in Appendix \ref{appendix_mf}.

\subsubsection{Stochastic Approximation Procedure (SAP)}
\label{sap}
Another estimator of the gradients of a hybrid architecture could exploit better mixing rates (i.e., minimum number of steps before the Markov chain's distribution is close to its stationary distribution with respect to total variation distance) afforded by approximate maximum likelihood learning. The key idea behind this procedure is to maintain a set of multiple persistent Gibbs sampling chains in the background from which we may sample during each update of the hybrid architecture. The details of this procedure are described in Appendix \ref{appendix_sap}.

\section{Related Work}
\label{lit_review}
There have been a vast array of approaches to semi-supervised learning that deviate from the original pre-training breakthrough. Some leverage auxiliary models  to encourage learning of discriminative information earlier at various stages of the learning process \parencite{bengio_greedy_2007,zhang_supervised_2014,lee_deeply-supervised_2014}.  Others adopt a manifold perspective to learning and attempt to learn a representation that is robust to small variations in the input (and thus generalize better to unseen, unlabeled samples) either through a penalty on the model's Jacobian \parencite{rifai_manifold_2011} or through a special regularizer \parencite{weston_deep_2012}.

A simpler alternative is through Entropy Regularization, where a standard architecture, such as a drop-out rectifier network, is used in a self-training scheme, where the model's own generated proxy labels for unlabeled samples are then used in a weighted secondary gradient \parencite{lee_pseudo-label:_2013}.  Other hybrid learning approaches follow similar ideas of \parencite{lasserre_principled_2006}, including the training schemes proposed in \textcite{ororbia_deep_hybrid_2015a,ororbia_deep_hybrid_2015b}, which built on the initial ideas of \textcite{larochelle_classify_2008,larochelle_learning_2012}, and of which the schemes of \textcite{calandra_learning_2012,zhou_online_2012} were shown to be special cases. The hybrid learning framework for DHBM and DHDA models described in this paper follow in the spirit of the compound learning objectives described in those studies.  However, they do differ slight from their predecessors, the SBEN and the HSDA, in that they do not always make use of an additional pure discriminative gradient (such as in the case of the DHBM).

The DHBM shares some similarites to that of the ``stitched-together'' DBM of \textcite{salakhutdinov_deep_2009} and the MP-DBM \parencite{goodfellow_multi-prediction_2013}, which was originally proposed as another way to circumvent the greedy pre-training of DBM's using a back-propagation-based approach on an unfolded inference graph.  The key advantage of the MP-DBM is that it is capable of working on a variety of variational inference tasks beyond classification (i.e., input completion, classification with missing inputs, etc.). We note that combining our hybrid learning framework with a more advanced Boltzmann architecture like the MP-DBM could yield even more powerful semi-supervised models.  The deep generative models proposed \cite{kingma_semi-supervised_2014} also share similarity to our work especially in their use of a recognition network to make inference of latent states in such models fast and tractable.

The generality of our hybrid learning framework, we argue, extends far beyond only learning DHBM's and DHDA's, as indicated by the general presentation of Algorithm \ref{learning_framework}.  In fact, any model that is capable of learning either discriminatively or generatively may be employed, such as Sum-Product Networks \parencite{poon_sum-product_2011} or recently proposed back-propagation-free models like the Difference-Target Propagation network \parencite{lee_difference_2015}.  

Furthermore, general enhancements to gradient descent learning (i.e., momemtum, regularization, etc.) may also be incorporated into our procedure.  For simplicity, we only chose to incorporate a drop-out scheme \citet{hinton_improving_2012} to our learning procedure since our comparative base-line (the psuedo-labeled MLP) also employed one.  During learning, all this entails is applying random binary masks to the hidden layer statistics calculated from the recognition network and once again to calculated mean-field statistics (which will set various hidden units to 0 under a given probability).  For the DHDA, the same binary masks used during calculation of layer-wise statistics are applied to the error deltas computed during back-propagation (as in \citet{hinton_improving_2012}).  Prediction in both the DHBM and DHDA simply requires multiplying layer-wise statistics by the probability used in drop-out learning.

\section{Experimental Results}
\label{experiments}

\subsection{Online Learning Results}
\label{captcha}

\begin{figure}[tb]
\includegraphics[width=\textwidth]{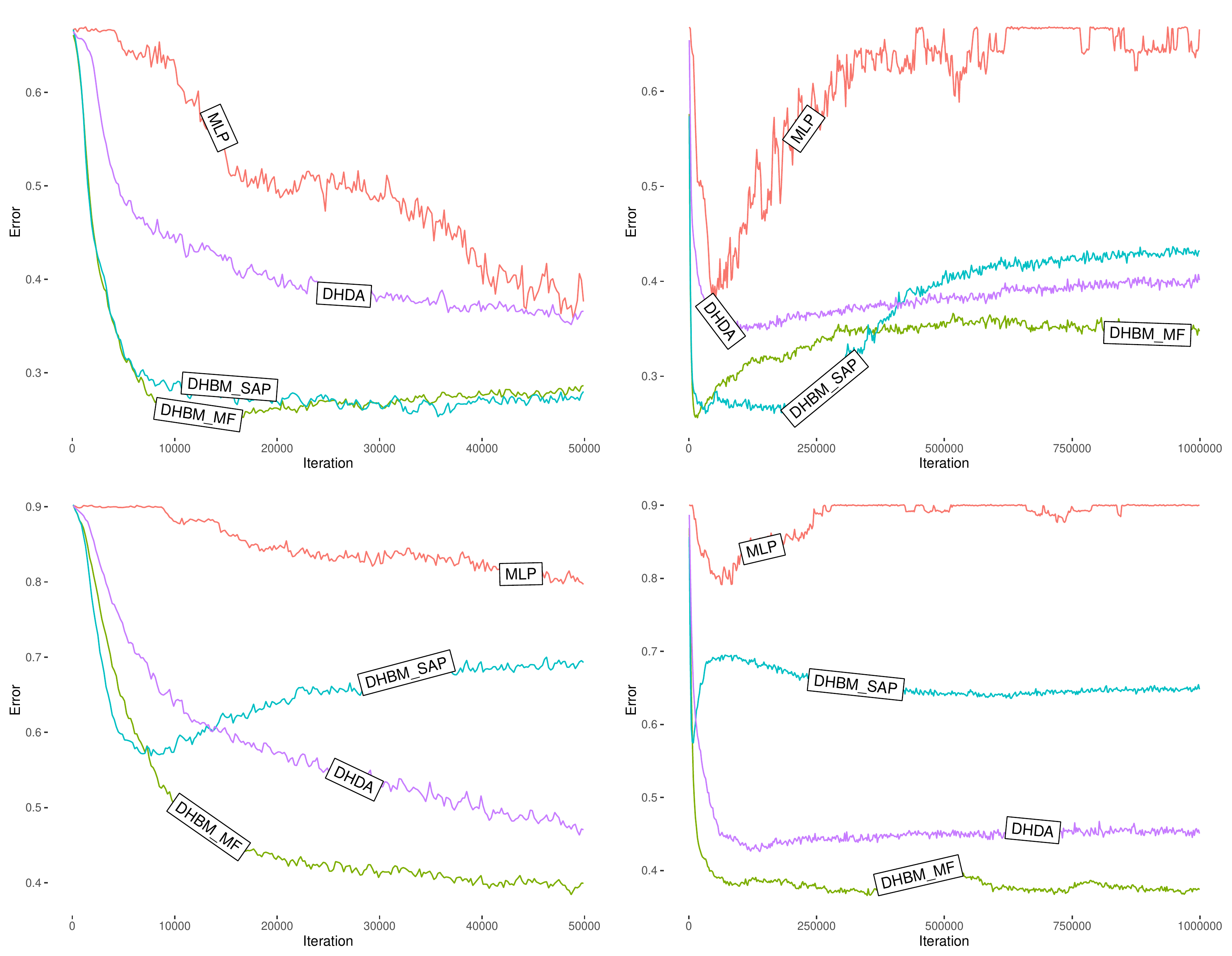}
\label{fig:byratio}  
  \caption{Online learning on the Waveform (top) and LED streams (bottom panels). Left panels zoom in to the first 50,000 iterations. 10\% labeled data (sampled uniformly 0-20\% in each trial).}
\end{figure}


We make use of the Massive Online Analysis framework (MOA, \cite{bifet_moa_2010}) to simulate the process of online learning, where the learner does not have access to a static, finite data-set of i.i.d. labeled samples but rather is presented with mini-batches of samples (we chose $N = 20$ samples at time) generated from a continuous non-stationary flow of non-i.i.d data.  We extended the evolving processes to the semi-supervised setting by enforcing that generated mini-batches are mixed in the sense that, on average, only 10\% of the samples were labeled at any time. To learn effectively, the learner must make use of these labeled samples as best as it can while also self-training from unlabeled samples.  

Specifically, the LED and Waveform streams were leveraged, in particular the versions that featured customizable feature concept drift and injection of feature noise to make learning more difficult over time.  The LED stream task entails predicting 1 of 10 possible digits on an LED panel given 24 noisy features (percentage of noise applied was set to $10\%$, and 4 features in the space experienced concept drift).  The Waveform task involves predicting 1 of 3 waveform types given 40 noisy real-valued attributes (these features were additionally normalized to lie in the range $[0,1]$, and 10 features in this space experienced concept drift).  For both data-streams, we investigate model performance over a 1,000,000 iteration sweep.

We compare 4 models:  1) the rectifier network trained via pseudo-labeled back-propagation of \textcite{lee_pseudo-label:_2013}, 2) a sigmoidal DHBM trained via MF-CD, 3) a sigmoidal DHBM trained via SAP (using 10 fantasy particles), and 4) a sigmoidal DHDA trained via MF-BP with corruption probability $p = 0.15$.  For both streams, all models consisted of complete-representation architectures (the LED stream used $24-24-24-24-24-10$ and the Waveform stream used $40-40-40-40-40-3$). All models used a drop-out probability of $p = 0.5$, a learning rate $\lambda = 0.051$, $\alpha = 1.0$, and $\beta = 0.1$ (we note that adaptive hyper-parameter schedules would be highly suitable to this learning setting and will be investigated in future work).


The results we report are 10-trial averages of each model's following the prequential method for evaluating statistical learning models for streaming scenarios.  Specifically, we make use of prequential error with a weighted forgetting factor $\alpha_{err} = 0.995$ (following the discussion of admissible values in \cite{gama_evaluating_2012}). This error metric is calculated (at time $i$) via the following equation:

\begin{equation}
\label{preq_error}
	 P_{\alpha}(i) = \frac{\sum^i_{k=1} \alpha^{i-k} L(y_k, \widehat{y}_k)}{\sum^i_{k=1} \alpha^{i-k}}\mbox{, with }0 \ll \alpha \leq 1 \mbox{.}
\end{equation}

\parencite{gama_evaluating_2012} showed that metrics with forgetting mechanisms were more appropriate for evolving data stream settings (as compared to simple predictive sequential error).  In particular, the one we use in this work is memoryless and thus advantageous to use in streaming settings over other alternatives such as sliding window prequential error. For each model, we show an error curve for the conditions when, on average, only 10\% of samples of a mini-batch at given time-step are labeled (Figure~\ref{fig:byratio}).

For both the LED and Waveform data-stream experiments, our proposed hybrid architectures consistently outperform the pseudo-labeled MLP.  However, we note that it is the \emph{DHBM\_MF} that performs the best (and not the \emph{DHBM\_SAP}), maintaining a consistently lower error in the face of both evolving data-streams (with the DHDA also performing reasonably well).  We speculate that the reason the SAP estimator does not help improve (and in fact, appears to hurt) performance is tied to the evolutionary nature of the input distribution itself.  SAP uses additional fantasy particles to better explore the DHBM's fantasy distribution and generally exhibit improved results over 1-step Contrastive Divergence in static data-set settings (where multiple epochs are possible).  However, since our input changes with time, it becomes more challenging to manipulate the model's fantasy distribution to match the input distribution.  The simpler mean-field CD approach, where the negative phase is simply 1 step away from the positive data-driven one, may simply facilitate an easier learning problem and thus allow for better adaptation of model parameters to handle changes in the input distribution.

Nonetheless, while all models ultimately experience fluctuations in error as the distributions change, we see that the \emph{DHBM\_MF} is even able to begin recovering in the Waveform experiment (roughly at around 60,000 iterations).  We observe that in the long-run, the MLP model ultimately performs quite poorly.  While the MLP is also semi-supervised, one reason behind its failure may be simply that its design is to make use of information useful for learning a discriminative mapping (or rather a conditional model $p(y|\mathbf{x})$). This is the case for both of its gradients (supervised and unsupervised). We argue that the key to our models' success is that they attempt to directly learn information about the input distribution itself in addition to conditional information.  The information afforded from including a generative perspective of the data seems to better equip hybrid architectures for input distributions that change with time than those without.

Additionally, it is likely that at certain time steps mini-batches presented to a learner are entirely unlabeled (as in our setting). It is here that semi-supervised models that also attempt to model the input distribution directly have an advantage over those that do not.  In this case, the pseudo-labeled MLP, which relies on the strength of its supervised gradient, is more likely to succumb to the problem most often associated with self-training schemes:  reinforcing incorrect predictions.

\subsection{Finite Data-Set Results}
\label{mnist}

\begin{table}[!t]
\caption{MNIST semi-supervised classification results (our results are  reported as 10-trial averages with standard error).}
\label{mnist_results}
\centering
\begin{tabular}{lr}
\multicolumn{1}{l}{}&\multicolumn{1}{c}{\begin{tabular}[x]{@{}c@{}}\textbf{Test Error}\\\end{tabular}}\tabularnewline
\hline
\textit{NN} & $25.81$\tabularnewline
\textit{SVM} & $23.44$\tabularnewline
\textit{CNN} & $22.98$ \tabularnewline
\textit{TSVM} & $16.81$ \tabularnewline
\textit{EMBEDNN \parencite{weston_deep_2012}} & $16.86$\tabularnewline
\textit{CAE \parencite{rifai_contractive_2011}} & $13.47$\tabularnewline
\textit{MTC \parencite{rifai_manifold_2011}} & $12.03$\tabularnewline
\textit{DROPNN \parencite{lee_pseudo-label:_2013}} & $21.89$\tabularnewline
\textit{DROPNN+PL \parencite{lee_pseudo-label:_2013}} & $16.15$\tabularnewline
\textit{DROPNN+PL+DAE \parencite{lee_pseudo-label:_2013}} & $10.49$\tabularnewline
\textit{3-DHBM} & $15.80 \pm 0.9$\tabularnewline
\textit{3-DHDA} & $21.24 \pm 0.6$\tabularnewline
\hline
\end{tabular}
\end{table}

We evaluate our proposed architectures on the MNIST data-set to gauge their viability for \emph{offline} semi-supervised learning.  The details of the experimental set-up can be found in Appendix \ref{mnist_setup}.  In Table \ref{mnist_results}~\footnote{For purely supervised models, \emph{NN} stands for a single-hidden layer neural network, \emph{SVM} stands for support sector machine, \emph{CNN} stands for convolutional neural network. For semi-supervised models, \emph{TSVM} is a transductive support vector machine, \emph{EMBEDNN} is deep neural network trained with an unsupervised embedding criterion, \emph{CAE} is a contractive auto-encoder, \emph{MTC} is the Manifold Tangent Classifier, and \emph{DROPNN} stands for a drop-out deep neural network (where \emph{PL} means pseudo-labeled and \emph{DAE} means with stacked denoising autoencoder pre-training.}, we observe that the \emph{3}-DHBM is competitive with several state-of-the-art models, notably the \emph{EMBEDNN} and \emph{DROPNN+PL}.  However, while the DHBM is unable to outperform the \emph{DROPNN+PL+DAE}, we note that that state-of-the-art method uses pre-training in order to obtain its final performance boost.  Further work could combine our hybrid architectures with the discriminatively-tracked pre-training approach proposed in Section~\ref{learning_framework}.  Alternatively, the DHBM could benefit from an additional weighted discriminative gradient like that used in the \emph{Top-Down-Bottom-Up} algorithm of \cite{ororbia_deep_hybrid_2015b}.

We also note that our DHDA model does not perform so well on this benchmark, only beating out the architectures trained on the supervised subset.  This poorer performance is somewhat surprising. However, we attribute this to too coarse a search for hyper-parameters (i.e., DHDA is particularly sensitive to $\lambda$ and $\beta$ and is affected by the corruption probability and type).  Its performance could also be improved by employing an SAP-like framework similar to that of the DHBM.

In an additional (5-trial) experiment using the 20 NewsGroup text data-set,  the 3-DHDA ($39.45 \pm 0.1$ \% test error) outperformed the \emph{DROPNN+PL} ($44.39 \pm 0.4$ \% test error) and 3-DHBM ($44.67 \pm 0.6$ \% test error).  The details of the experimental set-up can be found in Appendix \ref{20news_setup}.

\section{Conclusions}
\label{conclusions}
We have presented two novel deep hybrid architectures where parameters are learned jointly.  These unified hybrid models, unlike their predecessors, circumvent the problem of shifting representations since the different layers of these models are optimized from a global perspective.  Furthermore, prediction takes advantage of classification information found at all abstraction levels of the model without resorting to a vertical aggregation of disjoint, layer-wise experts (like that of the SBEN and HSDA),

Experiments show that our proposed unified hybrid architectures are well-suited to tackling more difficult, online data-stream settings.  We also observe that our architectures compare to some state-of-the-art semi-supervised learning methods on finite data-sets, even though we note that the online and offline tasks are substantially different.  More importantly, the unified hybrid learning framework described in this paper can be used beyond learning DHBM's and DHDA's.  Rather, we argue that it is applicable to any multi-level neural architecture that can compute discriminative and generative gradients for parameters jointly.  This implies that future work could train new hybrid variants of the models.

\subsubsection*{Acknowledgments}
We would like to thank Hugo Larochelle for useful conversations that helped inform this work.  The shortcomings of this paper, however, are ours and ours alone.  This work was supported by NSF grant 1528409 to DR.

\bibliography{main}

\appendix
\appendixpage

\section{Mean-Field Contrastive Divergence \& Back-Propagation Details}
\label{appendix_mf}
When gradients for a DHBM are estimated this way, one is effectively combining approximate mean-field variational inference with a multi-level Contrastive Divergence approximation.  This process, when it was introduced for learning unsupervised generative models like the DBM, is known Mean-Field Contrastive Divergence (MF-CD, \cite{welling_new_2002}).  However, training a DHBM with MF-CD is a bit simpler than for a DBM, since like the SBEN, the DHBM hybrid architecture allows for immediate classification and thus facilitates tracking of learning progress through a discriminative objective (such as the negative log loss).  The parameter gradients for each layer-wise expert are computed via the Contrastive Divergence approximation, without sampling the probability vectors obtained in the block Gibbs-sampling procedure.  The explicit procedure for estimating this gradient is given in Algorithm \ref{mean_field_cd}.

\begin{algorithm}[t]
\begin{algorithmic}
\State \textbf{Input:} 1) $(y,\mathbf{x})$ mini-batch of \emph{N} samples, 2) $Q^{rec}$ approximate factorial posterior for $(y,\mathbf{x})$, 3) $Q^{MF}$ mean-field factorial posterior for $(y,\mathbf{x})$, 3) initial model parameters \(\Theta^m = \{\Theta^m_1, \Theta^m_2,...,\Theta^m_L\}\), and 4) specialized hyper-parameters $\Xi$.

\Function{calcParamGradients}{$(y,\mathbf{x}), \hat{y}, Q^{rec}, Q^{MF}, \Theta^{m}, \Xi$}
    \State $l \gets 1$
    \While{$l \leq L$}
    	\LineComment{Gather positive phase statistics at \emph{l}}
    	\State $\mathbf{h}^{+}_l \gets Q^{rec}_l$
    	\If {$l > 1$}
        	\State $\mathbf{v}^{+}_l \gets Q^{rec}_l$
        \Else
 			\State $\mathbf{v}^{+}_l \gets \mathbf{x}$
        \EndIf
        \LineComment{Gather negative phase statistics at \emph{l}}
        \State $(\mathbf{v}^{-}_l,\mathbf{h}^{-}_l) \gets Q^{MF}_l$
        \LineComment{Calculate parameter gradients at \emph{l} via Contrastive Divergence}
        \State $\bigtriangledown^{W}_l \gets (<\mathbf{h}^{+}_l (\mathbf{v}^{+})^T>_N - <\mathbf{h}^{-}_l (\mathbf{v}^{-})^T>_N)$
        \State $\bigtriangledown^{U}_l \gets (<\mathbf{h}^{+}_l (\mathbf{e}_y)^T>_N - <(\mathbf{h}^{-}_l (\mathbf{e}_{\hat{y}})^T>_N)$
        \State $\bigtriangledown^m_l \gets (\bigtriangledown^{W}_l, \bigtriangledown^{U}_l)$
    \EndWhile
	\State \Return $\bigtriangledown^{m}$ \Comment{$\bigtriangledown^{m} = (\bigtriangledown^{m}_1,\bigtriangledown^{m}_1,...,\bigtriangledown^{m}_L)$}
\EndFunction

\end{algorithmic}
\caption{The estimator for calculating gradients using mean-field Contrastive Divergence for the DHBM.  Note that $\mathbf{e}_y$ is the vector representation of $y$ and $<>_N$ denotes calculating an expectation over \emph{N} samples.}
\label{mean_field_cd}
\end{algorithm}

In contrast, when this set of statistics is applied to a DHDA, one combines the mean-field variational inference procedure with a multi-level back-propagation of errors procedure.  Since each layer of the DHDA is a hybrid Denoising autoencoder (hDA), one may leverage the view in \textcite{ororbia_deep_hybrid_2015a} that such a building block is fusion of an encoder-decoder model with a single hidden-layer MLP.  The resulting component model may be trained via an unsupervised and supervised generative back-propagation procedure using a differentiable loss function such as reconstruction cross-entropy (or a quadratic loss).  The steps for calculating these gradient estimators using the mean-field statistics is shown in Algorithm \ref{mean_field_bp}.

\begin{algorithm}[t]
\begin{algorithmic}
\State \textbf{Input:} 1) $(y,\mathbf{x})$ mini-batch of \emph{N} samples, 2) $Q^{rec}$ approximate factorial posterior for $(y,\mathbf{x})$, 3) $Q^{MF}$ mean-field factorial posterior for $(y,\mathbf{x})$, 3) initial model parameters \(\Theta^m = \{\Theta^m_1, \Theta^m_2,...,\Theta^m_L\}\), and 4) specialized hyper-parameters $\Xi$.

\Function{calcParamGradients}{$(y,\mathbf{x}), \hat{y}, Q^{rec}, Q^{MF}, \Theta^{m}, \Xi$}
    \State $l \gets 1$, $\mathbf{\xi}^{out} \gets softmax^\prime(\hat{y}) \cdot -(y / \hat{y})$ \Comment{Calculate derivative of negative log-loss cost}
    \While{$l \leq L$}
    	\LineComment{Gather positive phase statistics at \emph{l}}
    	\State $\mathbf{h}^{+}_l \gets Q^{rec}_l$
    	\If {$l > 1$}
        	\State $\mathbf{v}^{+}_l \gets Q^{rec}_l$
        \Else
 			\State $\mathbf{v}^{+}_l \gets \mathbf{x}$
        \EndIf
        \LineComment{Gather negative phase statistics at \emph{l}}
        \State $(\mathbf{v}^{-}_l,\mathbf{h}^{-}_l) \gets Q^{MF}_l$
        \State $\widehat{\mathbf{z}}^{\mathbf{v}^{-}_l}_l \gets \mathbf{v}^{-}_l$, $\widehat{\mathbf{z}}^{\mathbf{h}^{-}_l}_l \gets \mathbf{h}^{-}_l$ \Comment{Get linear pre-activations for $\mathbf{v}^{-}_l$ \& $\mathbf{h}^{-}_l$}
		\State $\mathbf{\xi}^{recon}_{l} \gets \Call{derivReconLoss}{\mathbf{v}^{-}_l,\mathbf{v}^{+}_l}$, $\mathbf{\xi}^{recon}_l \gets \mathbf{\xi}^{recon}_l \cdot \phi^{\prime}(\widehat{\mathbf{z}}^{\mathbf{v}^{-}_l}_l)$ \Comment{Derivative of reconstruction loss at \emph{l}.}
        \State $\mathbf{\xi}^{hid}_{l} \gets (\Theta^{m}_l:W) \mathbf{\xi}^{recon}_{l}$ \Comment{Propagate error signal back to hiddens}
        \State $\mathbf{\xi}^{hid}_l \gets \mathbf{\xi}^{hid}_l \cdot \phi^{\prime}(\widehat{\mathbf{z}}^{\mathbf{h}^{-}_l}_l)$ \Comment{Compute error derivatives with respect to hiddens}        
        \State $\bigtriangledown^{W}_l \gets (\mathbf{\xi}^{hid}_l \mathbf{v}^{-}_l) + (\mathbf{\xi}^{recon}_l \mathbf{h}^{-}_l)$ \Comment{Compute 1st part of gradient for \emph{W}}
        \State $\mathbf{\xi}^{hid}_{l} \gets (\Theta^{m}_l:U) \mathbf{\xi}^{out}$ \Comment{Propagate output error signal back to hiddens}        
        \State $\bigtriangledown^{W}_l \gets \bigtriangledown^{W}_l + (\mathbf{\xi}^{hid}_{l} \mathbf{v}^{-}_l)$ \Comment{Compute 2nd part of gradient for \emph{W}}        
        \State $\bigtriangledown^{U}_l \gets \mathbf{h}^{-}_l (\mathbf{\xi}^{out})^T$ \Comment{Compute gradient for \emph{U}}
        \State $\bigtriangledown^m_l \gets (\bigtriangledown^{W}_l, \bigtriangledown^{U}_l)$
    \EndWhile
	\State \Return $\bigtriangledown^{m}$ \Comment{$\bigtriangledown^{m} = (\bigtriangledown^{m}_1,\bigtriangledown^{m}_1,...,\bigtriangledown^{m}_L)$}
\EndFunction
    
\end{algorithmic}
\caption{The estimator for calculating gradients using back-propagation of errors for the DHDA. Note that ``$\cdot$'' indicates a Hadamard  product, $\xi$ is an error signal vector, the prime superscript indicates a derivative (i.e., $\phi^\prime(v)$ means first derivative of activation function $\phi(v)$).  The symbol $\Theta:W$ denotes an access to element $W$ contained in $\Theta$.}
\label{mean_field_bp}
\end{algorithm}

\section{Stochastic Approximation Procedure Details}
\label{appendix_sap}
The Stochastic Approximation Procedure specifically requires maintaining a set of persistent Markov Chains (randomly initialized), or set of $M$ fantasy particles $\mathbf{\mathit{X}}_t = \{\mathbf{x}_{t,1},...,\mathbf{x}_{t,M}\}$, from which we calculate an average over.  From an implementation perspective, each time we make a call to update the hybrid model's parameters, we sample a new state $\mathbf{x}_{t+1}$ given $\mathbf{x}_t$ via a transition operator $\mathcal{T}_{\Theta_t}(\mathbf{x}_{t+1} \gets \mathbf{x}_t)$, of which, like the DBM in \textcite{salakhutdinov_efficient_2010}, we use Gibbs sampling.  Maintaining a set of persistent Gibbs chains facilitates better mixing during the MCMC learning procedure and better exploration of the model's energy landscape. More importantly, as we take a gradient step to obtain $\Theta_{t+1}$ using a point estimate of the model's intractable expectation at sample $\mathbf{x}_{t+1}$, we obtain a better estimate for the gradient of the final hybrid model.

Constructing an SAP for the unified learning framework defined by Algorithm \ref{learning_framework} entails no further work beyond implementing a multi-level block Gibbs sampler for each of the $M$ fantasy particles used for learning.  The explicit procedure is shown in Algorithm \ref{mean_field_sap}.

\begin{algorithm}[t]
\begin{algorithmic}
\State \textbf{Input:} 1) $(y,\mathbf{x})$ mini-batch of \emph{N} samples, 2) $Q^{rec}$ approximate factorial posterior for $(y,\mathbf{x})$, 3) $Q^{MF}$ mean-field factorial posterior for $(y,\mathbf{x})$, 3) initial model parameters \(\Theta^m = \{\Theta^m_1, \Theta^m_2,...,\Theta^m_L\}\), and 4) specialized hyper-parameters $\Xi$.

\Function{calcParamGradients}{$(y,\mathbf{x}), \hat{y}, Q^{rec}, Q^{MF}, \Theta^{m}, \Xi$}
    \State $l \gets 1$
    \While{$l \leq L$}
    	\LineComment{Gather positive phase statistics at \emph{l}}
    	\State $\mathbf{h}^{+}_l \gets Q^{rec}_l$
    	\If {$l > 1$}
        	\State $\mathbf{v}^{+}_l \gets Q^{rec}_l$
        \Else
 			\State $\mathbf{v}^{+}_l \gets \mathbf{x}$
        \EndIf
        \LineComment{Gather fantasy particle samples at \emph{l}}
        \For{each particle $m = 1$ to $M$}
        	\State Sample $(\tilde{\mathbf{v}}^{t+1,m}_{l},\tilde{\mathbf{h}}^{t+1,m}_{l})$ given $(\tilde{\mathbf{v}}^{t,m}_{l},\tilde{\mathbf{h}}^{t,m}_{l})$ via block Gibbs sampling.
        \EndFor
        \LineComment{Calculate parameter gradients at \emph{l} via persistent Contrastive Divergence}
        \State $\bigtriangledown^{W}_l \gets (<\mathbf{h}^{+}_l (\mathbf{v}^{+}_l)^T>_N - <\tilde{\mathbf{h}}^{t+1,m}_{l} (\tilde{\mathbf{v}}^{t+1,m}_{l})^T>_M)$
        \State $\bigtriangledown^{U}_l \gets (\mathbf{h}^{+}_l (<\mathbf{e}_y)^T>_N - <\tilde{\mathbf{h}}^{t+1,m}_{l} (\mathbf{e}_{\hat{y}})^T>_M)$
        \State $\bigtriangledown^m_l \gets (\bigtriangledown^{W}_l, \bigtriangledown^{U}_l)$
    \EndWhile
	\State \Return $\bigtriangledown^{m}$ \Comment{$\bigtriangledown^{m} = (\bigtriangledown^{m}_1,\bigtriangledown^{m}_1,...,\bigtriangledown^{m}_L)$}
\EndFunction
\end{algorithmic}
\caption{The estimator for calculating gradients using Stochastic Maximum Likelihood for the DHBM (and possibly the DHDA).}
\label{mean_field_sap}
\end{algorithm}

\section{Experimental Set-Up Details}
\label{exp_setup}

\subsection{MNIST}
\label{mnist_setup}

To evaluate the viability of our proposed hybrid architectures, we investigate their performance on the well-known MNIST benchmark.  However, since our models are designed for the semi-supervised setting, we make use of a similar experimental setting to \textcite{lee_pseudo-label:_2013}, which entails only using a small subset of the original 60,000 training sample set as a labeled training set and with the rest treated as purely unlabeled data points.  We separate out 1000 unique samples for validation (i.e., to perform the necessary model selection for finite data-set settings).  We ensure that there is an equal or fairly representative portion of each class variable in the training and validation subsets.  The MNIST data-set contains 28 x 28 images with pixel feature gray-scale feature values in the range of $[0,255]$ of which we normalized to the range of $[0,1]$.  We use these normalized real-valued raw features as direct input to our models.

Model selection was performed using a coarse grid search, where the hyper-parameters $\lambda$, in the range $[0.05,0.11]$, and $\beta$ in the range $[0.35,0.9]$, were key values to explore that affected generalization performance the most.  We employed an annealing schedule for $\beta_f$,  following the formula:

\begin{equation}
\label{beta_anneal}
    \beta(t)=\left\{
                \begin{array}{ll}
                  0, & \text{if $t < T_1$}\\
                  \frac{t - T_1}{T_2 - T_1} \beta_f, & \text{if $T_1 < t < T_2$}\\
                  \beta_f, & \text{if $T_2 < t$}
                \end{array}
              \right.
\end{equation}

\noindent
where $T$ denotes a ``labeled epoch'' or full pass through the labeled subset.  We set $T_1 = 3$ and $T_2 = 300$ for our experiments.

For the stochastic approximation procedure used to learn the DHBM, we made use of $M = 10$ fantasy particles, and for the MF-BP used to train the DHDA, we used a corruption probability of $p = 0.2$.  In calculating gradient estimates, we used mini-batches of 10 samples each iteration and did not decay the learning rate for any model.  Model architectures were kept to complete representations of the 3 latent layer form: $784-784-784-784-10$. For both the DHDA and the DHBM, we employed a Drop-Out scheme \parencite{hinton_improving_2012}, with probability of $p = 0.5$ for randomly masking out latent variables to better protect against overfitting of the data.  We only applied drop-out to the latent variables of the model, though note dropping out input variables may also improve performance yet further on the MNIST data-set.  Models were trained for a full 6 epochs through the entire training set (where a full epoch includes the set of all labeled and unlabeled samples).

\subsection{20 NewsGroups Set-Up}
\label{20news_setup}
We also investigate the performance of our model's on the 20-NewsGroup text classification data-set.  We opted to use the time-split version, which contains a training set with approximately 10000 document samples and a separate test set 8000 document samples.  In regards to pre-processing of the text, we removed stop-words, applied basic stemming, and removed numerics.  To create the final low-level representation of the data, we used only the 2000 most frequently occurring terms after pre-processing and create a binary occurrence vector for each document.  There are 20 class targets, each a different topic of discussion of the text.

All models in this experiment were chosen to use 2 hidden layers totaling 1200 variables.  We compare a 2-DHBM and 2-DHDA against 2 layer rectifier drop-out network trained via pseudo-labeled back-propagation.  For the rectifier network, we use an architecture of $2000-600-600-20$.  For the 2-DHBM, we use logistic sigmoid activation functions and the same architecture as the rectifier network but do not use drop-out (as we found it worsened model performance on this data-set).  For the 2-DHDA, we use an drop-out architecture of $2000-650-550-20$ with rectifier activation functions (and thus use a quadratic cost as the objective function for the mean-field back-propagation sub-routine) with denoising corruption set to $p = 0.2$.  All models had their learning rate searched in the interval $[0.01,0.1]$ and used an annealing schedule for $\beta$ with $T_1 = 3$ and $T_2 = 600$, where values for $\beta$ were searched in $[0.35,0.9]$.

\end{document}